\documentclass[runningheads]{llncs}
\usepackage{graphicx}
\usepackage{amsmath}
\usepackage{multicol}
\usepackage{multirow}
\usepackage{amssymb}
\usepackage{amsfonts}
\usepackage{booktabs}
\usepackage{siunitx}
\usepackage{hyperref}
\usepackage{marvosym}
\usepackage{graphicx}
\usepackage{float}
\usepackage{subfigure}
\usepackage{xcolor}

\usepackage{color}

\begin{document}
\title{Achieving Fairness Through Channel Pruning for Dermatological Disease Diagnosis}
\titlerunning{Achieving Dermatological Disease Diagnosis Fairness}
%
\author{Qingpeng Kong\inst{1}
\and Ching-Hao Chiu\inst{2} 
\and Dewen Zeng\inst{1} 
\and Yu-Jen Chen\inst{2} 
\and Tsung-Yi Ho\inst{3}
\and Jingtong hu\inst{4}
\and Yiyu Shi\inst{1}$^{(\textrm{\Letter})}$  
}
\authorrunning{Q. Kong et al.}
\institute{Department of Computer Science and Engineering, University of Notre Dame, Notre Dame, IN, USA \\
\email{\{qkong2, dzeng2, yshi4\}@nd.edu} \\
\and Department of Computer Science, National Tsing Hua University, Taiwan\\
\email{gwjh101708@gapp.nthu.edu.tw, chenjuzen@gmail.com} \\
\and Department of Computer Science and Engineering, The Chinese University of Hong Kong, Hong Kong 
\email{tyho@cse.cuhk.edu.hk}
\and Department of Electrical and Computer Engineering, University of Pittsburgh, Pittsburgh, PA, USA 
\email{jthu@pitt.edu}
}

\maketitle              
\begin{abstract}
Numerous studies have revealed that deep learning-based medical image classification models may exhibit bias towards specific demographic attributes, such as race, gender, and age. Existing bias mitigation methods often achieve high level of fairness at the cost of significant accuracy degradation. In response to this challenge, we propose an innovative and adaptable Soft Nearest Neighbor Loss-based channel pruning framework, which achieves fairness through channel pruning. Traditionally, channel pruning is utilized to accelerate neural network inference. However, our work demonstrates that pruning can also be a potent tool for achieving fairness. Our key insight is that different channels in a layer contribute differently to the accuracy of different groups. By selectively pruning critical channels that lead to the accuracy difference between the privileged and unprivileged groups, we can effectively improve fairness without sacrificing accuracy significantly. Experiments conducted on two skin lesion diagnosis datasets across multiple sensitive attributes validate the effectiveness of our method in achieving state-of-the-art trade-off between accuracy and fairness. Our code is available at \href{https://github.com/Kqp1227/Sensitive-Channel-Pruning}{
\textcolor{magenta}{https://github.com/Kqp1227/Sensitive-Channel-Pruning}}. 

\end{abstract}

\section{Introduction}
In AI-powered medical image analysis, deep neural networks (DNNs) are adept at extracting vital statistical details like colors and textures from the provided training data. This data-centric approach enables the network to grasp task-specific characteristics, thereby enhancing accuracy in the intended task. Nonetheless, in the pursuit of optimal accuracy, the network might rely on certain data attributes present in some instances but not in others, potentially leading to biases against specific demographics, such as skin tone or gender.
For example, in dermatological disease diagnosis, DNNs tend to exhibit different 
prediction accuracy across patients of different skin colors or genders \cite{groh2021evaluating,kinyanjui2020fairness,seyyed2021underdiagnosis,tschandl2018ham10000}. 

Various methods have been proposed to mitigate bias in machine learning models. These studies can be classified based on the stage when the debiasing techniques are applied: pre-processing, in-processing, and post-processing. Pre-processing methods \cite{kamiran2012data,wang2022fairness,xu2023fairadabn,Yang2023An} are deployed to mitigate biases within the dataset, thus fostering fairness. In-processing methods \cite{CHIU2024103188,deng2023fairness,fan2021fairness,roh2020fr,Wan2023In-Processing,wang2021directional,yao2022improving} typically involve the development of distinct network architectures to minimize fairness discrepancies. As for post-processing methods \cite{friedler2019comparative,hardt2016equality,Yang2023An}, calibration is done at inference time by considering both the model's prediction and the sensitive attribute as input. While these methods can enhance fairness and mitigate bias, they often struggle to uphold high accuracy levels while achieving great fairness improvement. 

To attain better accuracy-fairness trade-off, we introduce a novel framework that leverages channel pruning, which is orthogonal to and can be applied in conjunction with any existing bias mitigation method. Our approach is based on the observation 
that the output channels of convolutional layers within a convolutional neural network exhibit varying sensitivities to different sensitive attributes, such as skin color or gender. Those channels that are sensitive to these attributes contribute greatly to the accuracy gap between privileged and unprivileged groups (defined as ``sensitive channels'' in this paper), while those that are not would have little impact (defined as ``insensitive channels''). Conventionally, the channel pruning technique selects channels that are not important to accuracy and prunes them to accelerate DNN inference. We can make ``off-label'' use of this technique and instead prune the sensitive channels to promote feature diversity and diminish the impact of discriminatory features, thus mitigating bias. Specifically, we utilize the Soft Nearest Neighbor Loss (SNNL) \cite{Frosst2019AnalyzingAI} to measure the entanglement level between different sensitive attributes in the representation space, and identity the sensitive channels as those with feature maps exhibiting low SNNL values.  

Through extensive experimentation on different skin disease datasets and backbone models, we demonstrated that our framework can achieve the state-of-the-art trade-off between accuracy and fairness. We also demonstrated that it can be effectively applied to various existing bias mitigation frameworks to boost their trade-offs.  


    



\section{Method}


\begin{figure}[!ht]
    \centering
    \includegraphics[scale = 0.35]{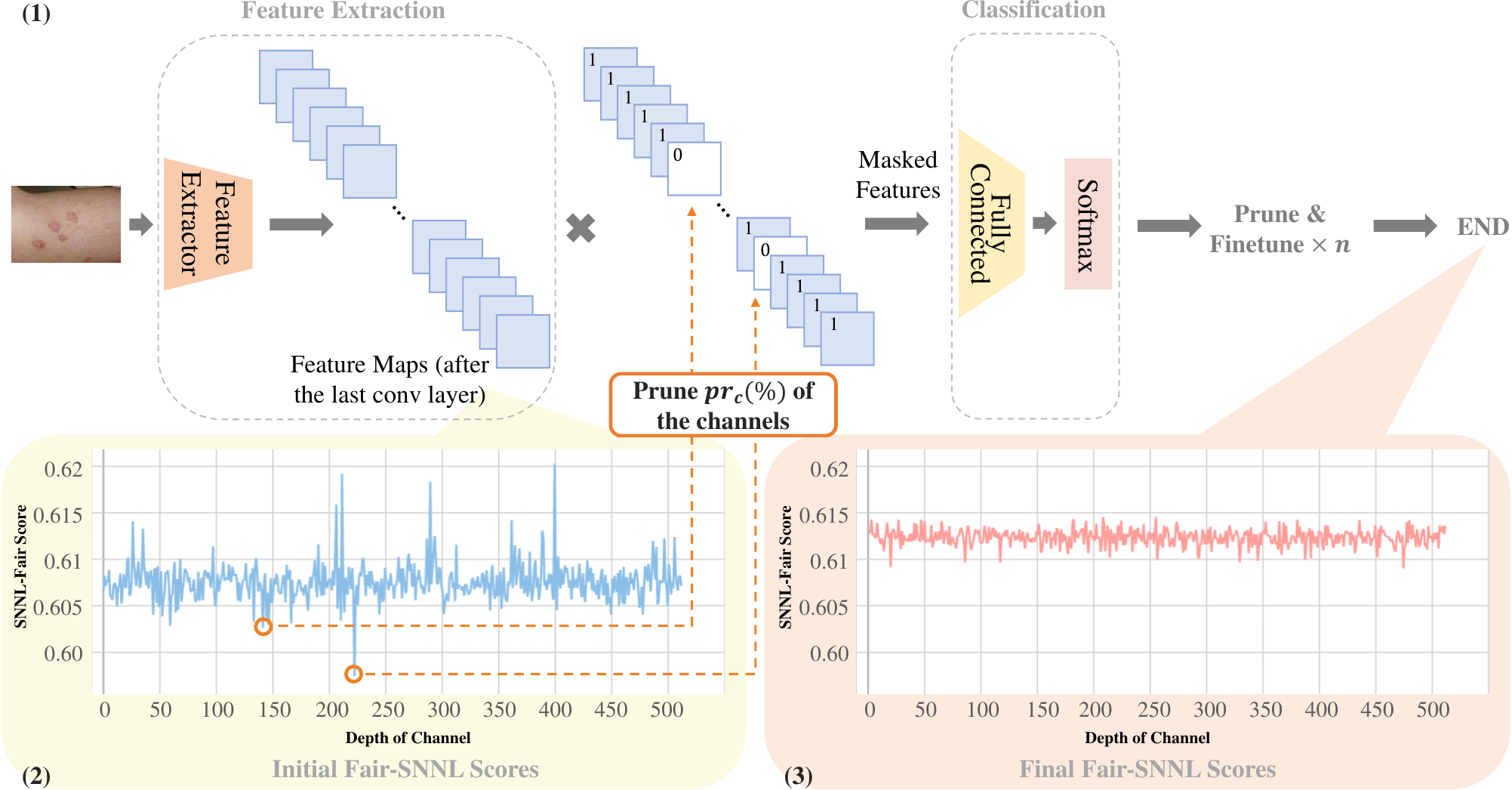}
    \caption{Illustration of (1) proposed channel pruning framework and (2) distribution of initial SNNL-Fair scores, generated using the Fitzpatrick-17k dataset and VGG-11 backbone and (3) distribution of the final SNNL-Fair scores after $n$ iterations of channel pruning. $pr_c$ represents channel pruning ratio in one iteration, and $n$ is the total number of pruning and fine-tuning iteration(s) needed under the stopping criteria.}
    \label{fig:workflow}
    \vspace*{-0.5cm}
\end{figure}
\subsection{Problem Formulation}
\label{sec:method_problem}
Consider a dataset $D = \{x_i, y_i, c_i\}, i=1,...,N$, where $x_i \in \mathbf x$ represents an input image ($\mathbf x$ represents all the input images), $y_i$ is the class label, and $c_i$ represents a sensitive attribute. Additionally, consider a pre-trained classification model $f_\theta(\cdot)$ with parameters $\theta$ that maps input $x_i$ to a predicted label $\hat{y}_i = f_\theta(x_i)$. 


Our goal is to selectively prune channels from a specific layer of the model $f_\theta(\cdot)$ that are sensitive to attribute $c_i$, thereby mitigating their influence on the overall fairness.
In this study, we only consider binary sensitive attributes, where $c_i \in \{0, 1\}$. Here, $c_i = 0$ corresponds to unprivileged samples, indicating discrimination by the model, while $c_i = 1$ corresponds to privileged samples.

\subsection{Channel Pruning for Fairness}
\subsubsection{Using SNNL to reflect entanglement between different sensitive groups.}
Conventionally, as mentioned in \cite{Frosst2019AnalyzingAI}, the Soft Nearest Neighbor Loss (SNNL) measures the entanglement state of features based on their class representation. A higher SNNL value indicates that the features belonging to different classes are intertwined, whereas a lower value suggests that the features are more separated by class. We can extend the concept and use it to evaluate the entanglement state of features based on their sensitive attributes. We define $\mathbf m$ for the feature maps extracted after the specific convolutional layer using the input images $\mathbf x$. For the $t$-th batch, let $b$ denote the batch size. $\mathbf m^{(t), k}=\{m^{(t), k}_1, m^{(t), k}_2, ..., m^{(t), k}_b\}$ denotes the feature maps extracted from the specific channel $k$ for the batch, and $\mathbf c^{(t)}=\{c^{(t)}_1, c^{(t)}_2, ..., c^{(t)}_b\}$ represents the corresponding sensitive labels. The modified SNNL $l_{sn}^{(t),k}$ to reflect sensitive attribute entanglement for channel $k$ of this batch $t$, with temperature $T$, can be calculated as follows:

\begin{equation} \label{eq:SNNL}
   l_{sn}^{(t), k}(\mathbf m^{(t), k}, \mathbf c^{(t)}, T)=-\frac{1}{b}\sum_{i \in 1..b} \log \left(\frac{\sum\limits_{\stackrel{j \in 1, ..., b }{\stackrel{j\neq i}{c_i^{(t)} = c_j^{(t)}}}} e^{-\frac{\Vert m_i^{(t), k}-m_j^{(t), k} \Vert ^2}{T}}}{\sum\limits_{\stackrel{p \in 1, ..., b}{p\neq i}} e^{-\frac{\Vert m_i^{(t), k}-m_p^{(t), k} \Vert ^2}{T}}} \right)
\end{equation}

Assume that the entire dataset forms a total of $n_b$ batches, 
then we will obtain a total of $n_b$ SNNL values $l_{sn}^{(1), k}, l_{sn}^{(2), k}, ..., l_{sn}^{(n_b), k}$. We then average 
these values to obtain a single value, defined as SNNL-Fair 
score $S^k$ for channel $k$, i.e., $S^k = \frac{1}{n_b}\sum_{t=1}^{n_b} l_{sn}^{(t), k}$.
For interested readers, a visual illustration of the SNNL-Fair calculation is included our supplementary materials.

\subsubsection{Using channel pruning to achieve fairness.}

In our methodology, we adopt an in-processing approach that entails pruning the sensitive channels. Even though our method can be applied to any layers, we choose to only prune the output channels from the last convolutional layer. This is because feature maps from deeper layers contain more detailed information about the input and thus providing better estimation of SNNL and more effective pruning. In the experiments we will present ablation studies to show that pruning the output channels of the last convolutional layer indeed presents the best opportunity for bias mitigation and accuracy preservation.  



As discussed, the SNNL-Fair scores reflect the entanglement state of different sensitive attributes in the feature maps of each channel. A smaller SNNL-Fair score for a channel indicates a stronger ability to differentiate between different sensitive attributes in its feature maps, potentially creating bias. As we can see from Fig.~\ref{fig:workflow}(2), using VGG-11 as backbone and Fitzpatrick-17k dataset as an example, there indeed exist certain channels with significantly lower SNNL-Fair scores than others (circled in orange), which means that different sensitive attributes are less entangled in these channels. To enhance fairness, we propose to prune them. Specifically, we sort the channels based on 
their SNNL-Fair scores, and prune the bottom $pr_c$ percent, where $pr_c$ is the channel pruning ratio. In the ablation study, 
we will study its impact on accuracy and fairness.  

Note that similar to existing channel pruning methods for 
inference speedup, we can adopt the iterative pruning for fairness enhancement. After each iteration of pruning, we fine-tune 
the pruned model on the training dataset to update the 
weights. 
The iterations stop either when the accuracy drop compared with the original unpruned model is above a threshold $th_{acc}$, or when the fairness improvement compared with the previous iteration is smaller than a threshold $th_{fair}$. 

We illustrate our framework in Fig.~\ref{fig:workflow}. Upon implementing our framework, we observed improvements in the overall SNNL-Fair scores, as illustrated in Fig.~\ref{fig:workflow}(3). The blue and red curves represent the SNNL-Fair score of each output channel in the last convolutional layer of VGG-11 before and after applying our framework, respectively. It is interesting to see that even though we only choose to prune the channels with lowest SNNL-Fair socre, after several iterations of pruning and fine-tuning, the SNNL-Fair scores across all the channels have increased. As previously mentioned, higher SNNL-Fair scores suggest improved fairness, implying that the model's fairness has been enhanced.

\subsection{Pruning Recipe}
Our framework prunes a pre-trained model by the following steps: 

1. For each channel $k$ in the last convolutional layer, compute the SNNL-Fair Scores $S^k$ using the given training dataset $D$. 

2. Sort the channels in descending order of $S^k$.

3. Prune $pr_c$ percent of the channels with smallest SNNL-Fair scores, resulting in a new model $f_\theta'(\cdot)$.


4. Fine-tune the new model $f_\theta'(\cdot)$ on $D$.

5. Repeat steps 1-4 until the stopping criteria are met. 



\section{Experiments and Results}
\subsection{Experiments}
\subsubsection{Dataset and Models} \label{sec:dataset}

The proposed methods are evaluated on two dermatology datasets for disease classification: the Fitzpatrick-17k dataset\cite{groh2021evaluating} and the ISIC 2019 challenge dataset\cite{combalia2019bcn20000}. The Fitzpatrick-17k dataset comprises 16,577 images representing 114 different skin conditions. The dataset includes individuals with a range of skin tones, 
and we further group the skin tones into two categories: light skin and dark skin. For our analysis, we consider skin tones as the sensitive attribute, representing the situation of ``explicit'' attribute (observable from the images). 
Containing a diverse collection of 25,331 dermoscopic images, the ISIC 2019 dataset  offers resource for studying skin conditions across various 8 diagnostic categories. We consider gender as the sensitive attribute, representing the situation of ``implicit'' attribute (unobservable from the images).
We follow existing works and use VGG-11 \cite{simonyan2014very} as the backbone for Fitzpatrick-17k and ResNet18 \cite{he2016deep} for ISIC2019. This allows us to assess the adaptability of our framework across different backbones and various datasets. All the training settings were the same with those described in \cite{chiu2023toward,wu2022fairprune}.

A standard preprocessing step for both datasets involves resizing all the images to a uniform size of 128×128 pixels. To augment the data and improve generalization, we apply various techniques such as random horizontal flipping, vertical flipping, rotation, scaling, and autoaugment, consistent with \cite{chiu2023toward,wu2022fairprune}.

\subsubsection{Fairness Metrics} \label{sec:metrics}
We employ the multi-class equalized opportunity (\textit{Eopp}) and equalized odds (\textit{Eodd}) metrics \cite{hardt2016equality} to evaluate the fairness of our model. 
We follow the approach of \cite{wu2022fairprune} for calculating these metrics. 
As there is always trade-off between accuracy and fairness, 
to better compare different methods, we employ the \textit{FATE} metric proposed by \cite{xu2023fairadabn} to evaluate the overall performance of bias mitigation methods. A higher \textit{FATE} score indicates that the model reaches a better trade-off between fairness and accuracy. \cite{CHIU2024103188} It can be calculated as $FATE_{FC} = \frac{ACC_m - ACC_b}{ACC_b} - \lambda\frac{FC_m - FC_b}{FC_b}$, 
where $FC$ can use one of the values from \textit{Eopp0}, \textit{Eopp1}, or \textit{Eodd}. \textit{ACC} represents accuracy, which can be chosen from accuracy scores, including F1-score, Recall, and Precision. Here, we choose the F1-score for comparison. The subscripts $m$ and $b$ represent the bias mitigation and baseline models, respectively. The parameter $\lambda$ is used to modify the significance of fairness in the ultimate evaluation, and we set the $\lambda = 1.0$ following the same setting used in \cite{xu2023fairadabn}.

\renewcommand{\arraystretch}{2}
\begin{table*}[!ht]
\caption{Results of accuracy and fairness of various methods on the Fitzpatrick-17k  and VGG-11 backbone, using skin tone as the sensitive attribute. The dark skin is the privileged group. \textit{FATE} metrics are evaluated using the vanilla VGG-11 as the baseline. 
}\label{fitz_main_result}
\vspace{-0.5cm}
\begin{center}
\renewcommand{\arraystretch}{0.8}
\resizebox{\linewidth}{!}{
\begin{tabular}{ccccccccc}
\toprule
& & \multicolumn{3}{c}{Accuracy} & \multicolumn{3}{c}{Fairness}\\
\cmidrule{3-5}\cmidrule(l{2pt}r{2pt}){6-8}
Method & Skin Tone & Precision & Recall & F1-score & \textit{Eopp0}↓ / \textit{FATE}↑ & \textit{Eopp1}↓ / \textit{FATE}↑ & \textit{Eodd}↓ / \textit{FATE}↑ \\


\midrule
\multirow{4}{*}{VGG-11~\cite{simonyan2014very}} 
&Dark & 0.563 & 0.581 & 0.546 & \multirow{4}{*}{0.0013 / 0.0000} & \multirow{4}{*}{0.361 / 0.0000} & \multirow{4}{*}{0.182 / 0.0000} \\ 
&Light & 0.482 & 0.495 & 0.473 \\ 
&Avg.↑ & 0.523 & 0.538 & 0.510 \\ 
&Diff.↓ & 0.081 & 0.086 & 0.073 \\

\midrule
\multirow{4}{*}{HSIC~\cite{quadrianto2019discovering}} 
&Dark & 0.548 & 0.522 & 0.513  & \multirow{4}{*}{0.0013 / -0.0196} & \multirow{4}{*}{0.331 / 0.1235} & \multirow{4}{*}{0.166 / 0.1305} \\ 
&Light & 0.513 & 0.506 & 0.486 \\ 
&Avg.↑ & 0.530 & 0.515 & 0.500 \\ 
&Diff.↓ & 0.040 & 0.018 & 0.029 \\


\midrule
\multirow{4}{*}{MFD~\cite{jung2021fair}} 
&Dark & 0.514 & 0.545 & 0.503  & \multirow{4}{*}{\underline{0.0011} / 0.0950} & \multirow{4}{*}{0.334 / 0.0160} & \multirow{4}{*}{0.166 / 0.0291} \\ 
&Light & 0.489 & 0.469 & 0.457 \\ 
&Avg.↑ & 0.502 & 0.507 & 0.480 \\ 
&Diff.↓ & 0.025 & 0.076 & 0.046 \\

\midrule
\multirow{4}{*}{FairAdaBN~\cite{xu2023fairadabn}} 
&Dark & 0.544 & 0.541 & 0.524  & \multirow{4}{*}{0.0012 / 0.0583} & \multirow{4}{*}{0.341 / 0.0368} & \multirow{4}{*}{0.171 / 0.0418} \\ 
&Light & 0.484 & 0.509 & 0.476 \\ 
&Avg.↑ & 0.514 & 0.525 & 0.500 \\ 
&Diff.↓ & 0.033 & 0.060 & 0.048 \\

\midrule
&Dark & 0.567 & 0.519 & 0.507   & \multirow{4}{*}{\textbf{0.0008 / 0.3317}} & \multirow{4}{*}{0.330 / 0.0329} & \multirow{4}{*}{0.165 / 0.0405} \\ 
FairPrune~\cite{wu2022fairprune} &Light & 0.496 & 0.477 & 0.459 \\ 
($pr=35\%, \beta=0.33$) &Avg.↑ & 0.531 & 0.498 & 0.483 \\ 
&Diff.↓ & 0.071 & 0.042 & 0.048 \\

\midrule
\multirow{4}{*}{ME-FairPrune~\cite{chiu2023toward}} 
&Dark & 0.564 & 0.529 & 0.523  & \multirow{4}{*}{0.0012 / \underline{0.1005}} & \multirow{4}{*}{\underline{0.305} / \underline{0.1787}} & \multirow{4}{*}{\underline{0.152} / \underline{0.1884}} \\ 
&Light & 0.542 & 0.535 & 0.522 \\ 
&Avg.↑ & 0.553 & 0.532 & 0.522 \\ 
&Diff.↓ & 0.022 & 0.006 & 0.001 \\

\midrule
&Dark & 0.568 & 0.576 & 0.547   & \multirow{4}{*}{0.0012 / 0.0965} & \multirow{4}{*}{\textbf{0.278 / 0.2495}} & \multirow{4}{*}{\textbf{0.139 / 0.2559}} \\ 
SCP-FairPrune (Ours) &Light & 0.499 & 0.504 & 0.492 \\ 
($pr_c=2\%, n=3$) &Avg.↑ & 0.533 & 0.540 & 0.520 \\ 
&Diff.↓ & 0.069 & 0.073	& 0.055 \\

\bottomrule
\end{tabular}}
\end{center}
\end{table*}
\subsection{Comparison with State-of-the-art}
\label{Compare_SOTA}


We present a comprehensive comparison of our framework against various bias mitigation baselines~\cite{chiu2023toward,jung2021fair,lecun1989optimal,quadrianto2019discovering,simonyan2014very,tzeng2015simultaneous,wang2020towards,wu2022fairprune,xu2023fairadabn,zhang2018mitigating}. Due to space limit, we are only able to include the most recent baselines here, and a comprehensive comparison with several other baselines are included in the supplementary material. We denote our framework as SCP (\underline{S}NNL based \underline{C}hannel \underline{P}runing), and apply it to the model from state-of-the-art bias mitigation method FairPrune~\cite{wu2022fairprune}. The resulting model is named SCP-FairPrune. 
$n$ is the number of iterations undertaken by our framework before the stopping criteria are met, and $pr_c$ is the channel pruning ratio in each iteration (set to 2\%). \textbf{Bold} and \underline{Underline} fonts denote the best and the second-best performance, respectively.

\vspace{-0.5cm}
\subsubsection{Results on Fitzpatrick-17k Dataset} 
The results presented in Table~\ref{fitz_main_result} demonstrate the superior performance of our framework compared to other methods, as evidenced by the lowest \textit{Eopp1} and \textit{Eodd} scores. Notably, 
the \textit{FATE} metric of \textit{Eopp1} and \textit{Eodd} demonstrate enhancements of 39.7\% and 35.8\% over the last state-of-the-art method ME-FairPrune respectively, 
which proves its effectiveness in enhancing trade-off between fairness and accuracy. Note that this improvement is achieved within only three iterations of pruning and less than 10\% of the channels pruned in total. 

As our framework is orthogonal to existing bias mitigation methods, we also evaluate its effectiveness when being applied to other baselines, including the vanilla VGG-11 (i.e., SCP-VGG-11) and 
another bias mitigation method HSIC (i.e., SCP-HSIC). Results show that the SCP-VGG-11 and SCP-HSIC both achieve positive \textit{FATE} metrics over their respective baselines (details in the supplementary material). This demonstrates that our framework can improve the trade-off between accuracy and fairness on a wide range of methods.  

\subsubsection{Results on ISIC 2019 Dataset} 
With ResNet-18 as backbone, the results presented in Table \ref{isic_result} show that our framework still achieves the highest \textit{FATE} metric of \textit{Eopp1} and \textit{Eodd}, demonstrating enhancements of 22.0$\%$ and 33.9$\%$ over ME-FairPrune, respectively. 

\begin{table*}[!ht]
\caption{Results of accuracy and fairness of various methods on the ISIC 2019 and ResNet-18 backbone, using gender as the sensitive attribute. Female is the privileged group. \textit{FATE} metrics are evaluated using the vanilla ResNet-18 as the baseline. 
}\label{isic_result}
\vspace{-0.5cm}
\begin{center}
\renewcommand{\arraystretch}{0.8}
\resizebox{\linewidth}{!}{
\begin{tabular}{ccccccccc}
\toprule
& & \multicolumn{3}{c}{Accuracy} & \multicolumn{3}{c}{Fairness}\\
\cmidrule{3-5}\cmidrule(l{2pt}r{2pt}){6-8}
Method & Gender & Precision & Recall & F1-score & \textit{Eopp0}↓ / \textit{FATE}↑ & \textit{Eopp1}↓ / \textit{FATE}↑ & \textit{Eodd}↓ / \textit{FATE}↑ \\
\midrule
\multirow{4}{*}{ResNet-18~\cite{he2016deep}} 
&Female & 0.793 & 0.721 & 0.746 & \multirow{4}{*}{0.006 / 0.0000} & \multirow{4}{*}{0.044 / 0.0000} & \multirow{4}{*}{0.022 / 0.0000} \\ 
&Male & 0.731 & 0.725 & 0.723 \\ 
&Avg.↑ & 0.762 & 0.723 & 0.735 \\ 
&Diff.↓ & 0.063 & 0.004 & 0.023 \\




\midrule
\multirow{4}{*}{HSIC~\cite{quadrianto2019discovering}} 
&Female & 0.744 & 0.660 & 0.696  & \multirow{4}{*}{0.008 / 0.8194} & \multirow{4}{*}{0.042 / 0.0068} & \multirow{4}{*}{0.020 / 0.0559} \\ 
&Male & 0.718 & 0.697 & 0.705 \\ 
&Avg.↑ & 0.731 & 0.679 & 0.700 \\ 
&Diff.↓ & 0.026 & 0.037 & 0.009 \\

\midrule
\multirow{4}{*}{MFD~\cite{jung2021fair}} 
&Female & 0.770 & 0.697 & 0.726  & \multirow{4}{*}{\textbf{0.005} / \underline{0.9171}} & \multirow{4}{*}{0.051 / -0.1482} & \multirow{4}{*}{0.024 / -0.0758} \\ 
&Male & 0.772 & 0.726 & 0.744 \\ 
&Avg.↑ & 0.771 & 0.712 & 0.735 \\ 
&Diff.↓ & 0.002 & 0.029 & 0.018 \\

\midrule
\multirow{4}{*}{FairAdaBN~\cite{xu2023fairadabn}} 
&Female & 0.776 & 0.724 & 0.746  & \multirow{4}{*}{\textbf{0.005} / \textbf{0.9273}} & \multirow{4}{*}{0.038 / 0.1548} & \multirow{4}{*}{0.019 / 0.1586} \\ 
&Male & 0.758 & 0.728 & 0.739 \\ 
&Avg.↑ & 0.767 & 0.726 & 0.743 \\ 
&Diff.↓ & 0.008 & 0.004 & 0.007 \\

\midrule
&Female & 0.776 & 0.711 & 0.734   & \multirow{4}{*}{0.007 / 0.8729} & \multirow{4}{*}{0.026 / 0.4039} & \multirow{4}{*}{0.014 / 0.3617} \\ 
FairPrune~\cite{wu2022fairprune} &Male & 0.721 & 0.725 & 0.720 \\ 
($pr=35\%, \beta=0.33$) &Avg.↑ & 0.748 & 0.718 & 0.727 \\ 
&Diff.↓ & 0.055 & 0.014 & 0.014 \\

\midrule
\multirow{4}{*}{ME-FairPrune~\cite{chiu2023toward}} 
&Female & 0.770 & 0.723 & 0.742  & \multirow{4}{*}{\underline{0.006} / 0.9018} & \multirow{4}{*}{\underline{0.020} / \underline{0.5513}} & \multirow{4}{*}{\underline{0.010} / \underline{0.5533}} \\ 
&Male & 0.739 & 0.728 & 0.730 \\ 
&Avg.↑ & 0.755 & 0.725 & 0.736 \\ 
&Diff.↓ & 0.032 & 0.005 & 0.012 \\

\midrule
&Female & 0.787 & 0.701 & 0.736  & \multirow{4}{*}{\underline{0.006} / 0.9018} & \multirow{4}{*}{\textbf{0.015 / 0.6724}} & \multirow{4}{*}{\textbf{0.006 / 0.7411}} \\ 
SCP-FairPrune (Ours) &Male & 0.765 & 0.712 & 0.735 \\ 
($pr_c=2\%, n=3$) &Avg.↑ & 0.776 & 0.707 & 0.736 \\ 
&Diff.↓ & 0.022 & 0.012	& 0.001 \\

\bottomrule
\end{tabular}}
\end{center}
\end{table*}

\subsection{Ablation Study}
\subsubsection{Effect of Channel Pruning Ratio $pr_c$}
To explore how altering channel pruning ratio $pr_c$ affects accuracy and fairness, we employ our framework on a pre-trained VGG-11 model sourced from the Fitzpatrick-17k dataset and experiment with various ratios. The original model without pruning is used as the baseline for \textit{FATE} metric evaluation of $Eodd$. As depicted in Figure \ref{Fig.sub.1}, the model achieves the highest \textit{FATE} score when setting the pruning ratio to $2\%$. It is also interesting to note that the optimal results can be achieved by no more than three pruning iterations, demonstrating the efficiency of our proposed framework; running more-than-necessary pruning iterations will cause the performance to drop, showcasing the importance of using proper stopping criteria (Note that in this study and the study below, we allowed the pruning iterations to go beyond the stopping criteria to show the peaks). 
\subsubsection{Effect of Pruning Channels in Different Layers}
\begin{figure}[t]
\label{ablation}
\centering  
\subfigure[]{
\label{Fig.sub.1}
\includegraphics[scale = 0.23]{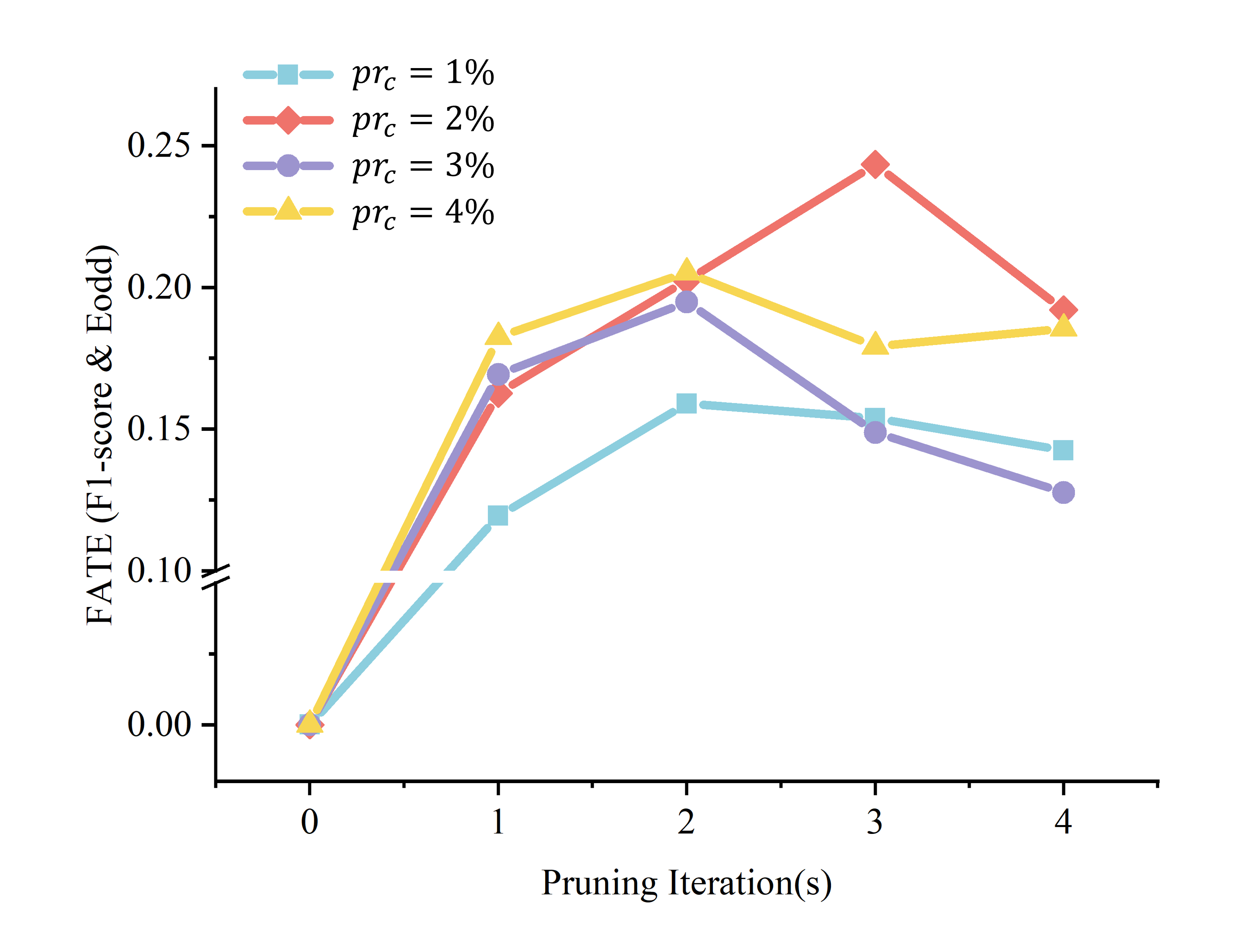}}\subfigure[]{
\label{Fig.sub.2}
\includegraphics[scale = 0.23]{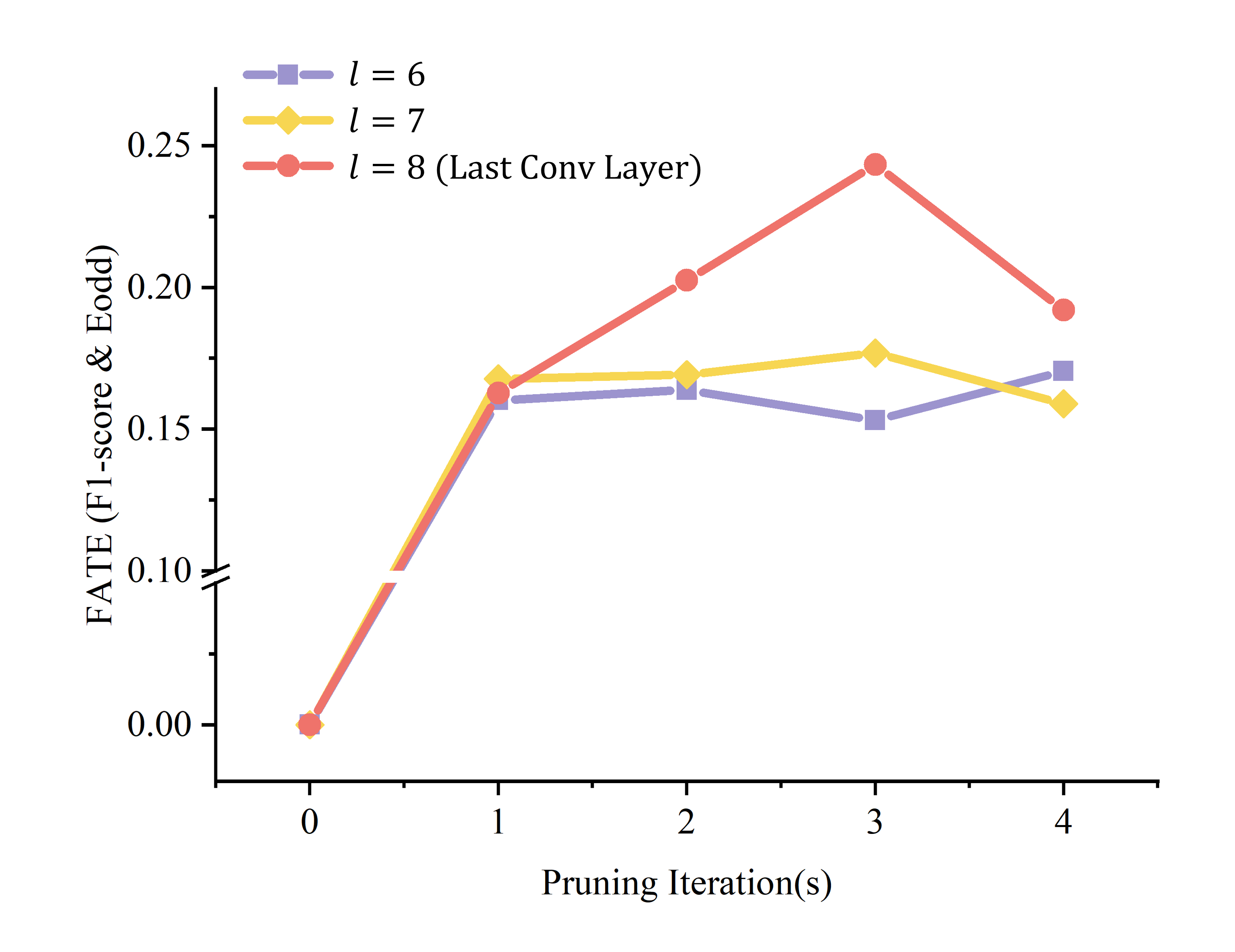}}
\caption{\textit{FATE} of $Eodd$ v.s. pruning iterations with (a) varying channel pruning ratio $pr_c$ and (b) varying layer from which the output channels are pruned, using VGG-11 on Fitzpatrick-17k dataset. }
\vspace{-0.5cm}
\end{figure}

The VGG-11 model features a total of 8 convolutional layers, and our framework prunes the output channels of the 8th (last) layer. We experimented with pruning the channels in other layers (index $l$) using a 2\% pruning ratio on the Fitzpatrick-17k dataset and recorded the resulting \textit{FATE} metrics of \textit{Eodd}. In Fig.~\ref{Fig.sub.2}, though pruning the channels in other layers can also improve fairness and accuracy trade-off, it is evident that working with the last convolutional layer yields the best result.

\section{Conclusion}

We tackle the challenge of declining fairness in convolutional neural networks by introducing a Channel Pruning framework based on SNNL (Soft Nearest Neighbor Loss). Our framework is versatile, compatible with a range of bias mitigation techniques, and selectively prunes the sensitive channels from specific convolutional layers. Experiment results underscore the effectiveness of our framework, showcasing superior trade-offs between accuracy and fairness compared to state-of-the-art methods across two dermatological disease datasets.


\newpage
{\small
\bibliographystyle{splncs04}
\bibliography{MICCAI}
}

%
%
%
%




\newpage
\setcounter{page}{1} 
\renewcommand\thesection{\Alph{section}}
\setcounter{section}{0}
\setcounter{theorem}{0}
\setcounter{definition}{0}
\setcounter{figure}{0}
\setcounter{table}{0}

\begin{figure}[!ht]
    \centering
    \includegraphics[scale = 0.3]{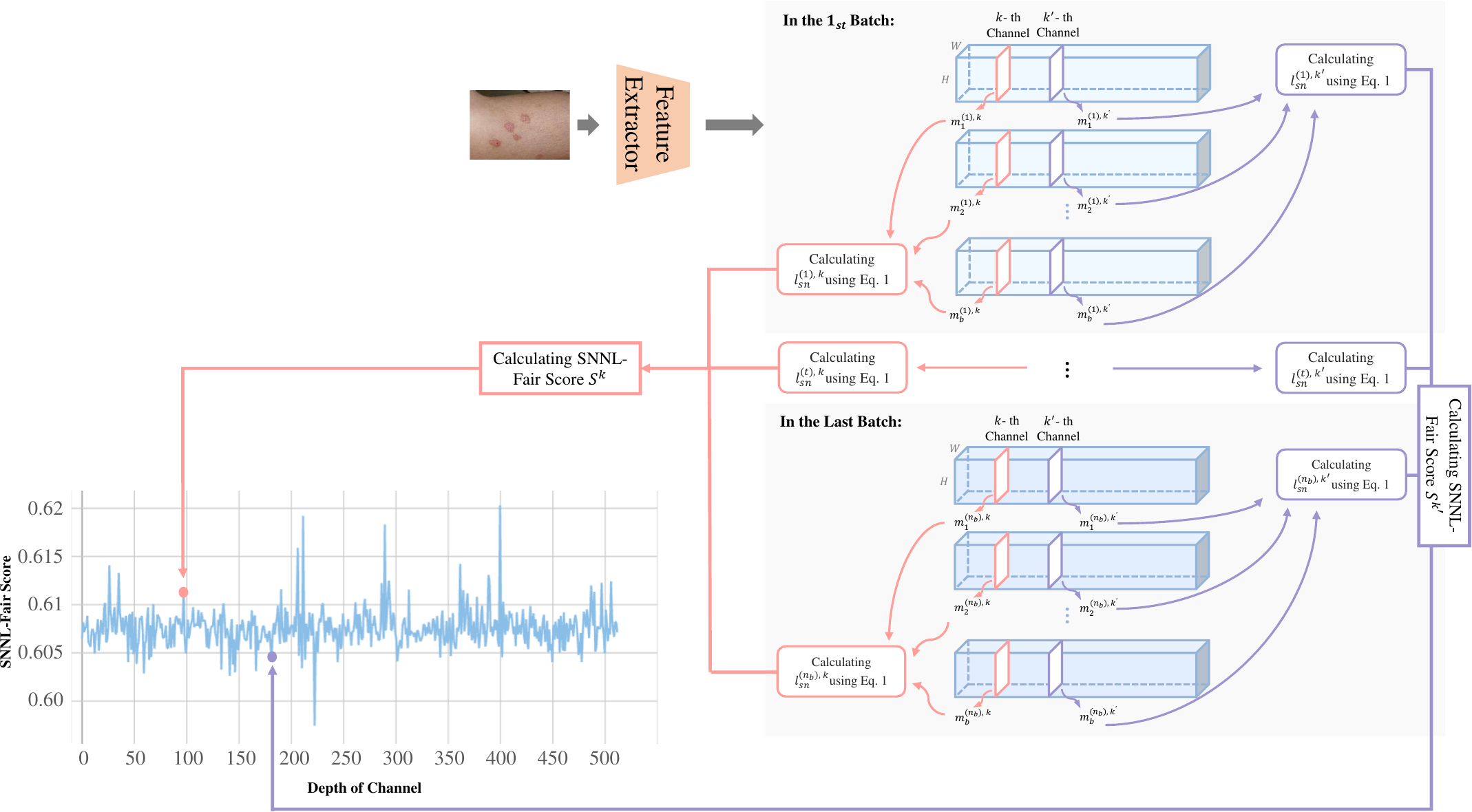}
    \caption{Illustration of SNNL-Fair metric calculation. Here, $t$ represents the batch index, $b$ represents the batch size, $k$ represent the depth of channel, $n_b$ represents total number of batches, and $m_b^{(t), k}$ represent the feature map at the $k$-th channel in the $t$-th batch. }
    \label{fig:workflow}
\end{figure}
\renewcommand{\arraystretch}{2}
\begin{table*}[!ht]
\caption{Additional results of accuracy and fairness on the Fitzpatrick-17k and VGG-11 backbone, using skin tone as the sensitive attribute. The dark skin is the privileged group. \textit{FATE} metrics are evaluated using the vanilla VGG-11 as the baseline. ($pr$ is the pruning ratio, $n$ is the pruning iteration(s), and $pr_c$ is the channel pruning ratio.) 
}\label{fitz_main_result}
\begin{center}
\renewcommand{\arraystretch}{0.8}
\resizebox{\linewidth}{!}{
\begin{tabular}{ccccccccc}
\toprule
& & \multicolumn{3}{c}{Accuracy} & \multicolumn{3}{c}{Fairness}\\
\cmidrule{3-5}\cmidrule(l{2pt}r{2pt}){6-8}
Method & Skin Tone & Precision & Recall & F1-score & \textit{Eopp0}↓ / \textit{FATE}↑ & \textit{Eopp1}↓ / \textit{FATE}↑ & \textit{Eodd}↓ / \textit{FATE}↑ \\



\midrule
\multirow{4}{*}{AdvConf~\cite{das2018mitigating}} 
&Dark & 0.506 & 0.562 & 0.506 & \multirow{4}{*}{\textbf{0.0011} / 0.0676} & \multirow{4}{*}{0.339 / -0.0253} & \multirow{4}{*}{0.169 / -0.0148} \\ 
&Light & 0.427 & 0.464 & 0.426 \\ 
&Avg.↑ & 0.467 & 0.513 & 0.466 \\ 
&Diff.↓ & 0.079 & 0.098 & 0.080 \\

\midrule
\multirow{4}{*}{AdvRef~\cite{tzeng2015simultaneous}} 
&Dark & 0.514 & 0.545 & 0.503  & \multirow{4}{*}{\textbf{0.0011} / \underline{0.0950}} & \multirow{4}{*}{\underline{0.334} / \underline{0.0160}} & \multirow{4}{*}{\underline{0.166} / \underline{0.0291}} \\ 
&Light & 0.489 & 0.469 & 0.457 \\ 
&Avg.↑ & 0.502 & 0.507 & 0.480 \\ 
&Diff.↓ & 0.025 & 0.076 & 0.046 \\

\midrule
\multirow{4}{*}{DomainIndep~\cite{wang2020towards}} 
&Dark & 0.547 & 0.567 & 0.532  & \multirow{4}{*}{0.0012 / 0.0416} & \multirow{4}{*}{0.344 / 0.0118} & \multirow{4}{*}{0.172 / 0.0197} \\ 
&Light & 0.455 & 0.480 & 0.451 \\ 
&Avg.↑ & 0.501 & 0.523 & 0.492 \\ 
&Diff.↓ & 0.025 & 0.076 & 0.046 \\

\midrule
&Dark & 0.557 & 0.570 & 0.536   & \multirow{4}{*}{\underline{0.0012} / 0.0691} & \multirow{4}{*}{0.360 / -0.0051} & \multirow{4}{*}{0.180 / 0.0031} \\ 
OBD~\cite{lecun1989optimal} &Light & 0.488 & 0.494 & 0.475 \\ 
(pr=35$\%$) &Avg.↑ & 0.523 & 0.532 & 0.506 \\ 
&Diff.↓ & 0.069 & 0.076 & 0.061 \\





\midrule
&Dark & 0.568 & 0.576 & 0.547   & \multirow{4}{*}{\underline{0.0012} / \textbf{0.0965}} & \multirow{4}{*}{\textbf{0.278 / 0.2495}} & \multirow{4}{*}{\textbf{0.139 / 0.2559}} \\ 
SCP-FairPrune (Ours) &Light & 0.499 & 0.504 & 0.492 \\ 
($pr_c=2\%, n=3$) &Avg.↑ & 0.533& 0.540 & 0.520 \\ 
&Diff.↓ & 0.069 & 0.073	& 0.055 \\

\bottomrule
\end{tabular}}
\end{center}
\end{table*}

\begin{table*}[!ht]
\caption{Additional results of accuracy and fairness on the ISIC 2019 dataset and ResNet-18 backbone, using gender as the sensitive attribute. Female is the privileged group. \textit{FATE} metrics are evaluated using the vanilla ResNet-18 as the baseline. ($n$ is the pruning iteration(s), and $pr_c$ is the channel pruning ratio.) 
}\label{isic_result}
\begin{center}
\renewcommand{\arraystretch}{0.8}
\resizebox{\linewidth}{!}{
\begin{tabular}{ccccccccc}
\toprule
& & \multicolumn{3}{c}{Accuracy} & \multicolumn{3}{c}{Fairness}\\
\cmidrule{3-5}\cmidrule(l{2pt}r{2pt}){6-8}
Method & Gender & Precision & Recall & F1-score & \textit{Eopp0}↓ / \textit{FATE}↑ & \textit{Eopp1}↓ / \textit{FATE}↑ & \textit{Eodd}↓ / \textit{FATE}↑ \\

\midrule
\multirow{4}{*}{AdvConf~\cite{das2018mitigating}} 
&Female & 0.755 & 0.738 & 0.741 & \multirow{4}{*}{0.008 / \underline{0.8684}} & \multirow{4}{*}{0.070 / -0.5748} & \multirow{4}{*}{0.037 / -0.6574} \\ 
&Male & 0.710 & 0.757 & 0.731 \\ 
&Avg.↑ & 0.733 & 0.747 & 0.736 \\ 
&Diff.↓ & 0.045 & 0.020 & 0.010 \\

\midrule
\multirow{4}{*}{AdvRef~\cite{tzeng2015simultaneous}} 
&Female & 0.778 & 0.683 & 0.716  & \multirow{4}{*}{\underline{0.007} / 0.8674} & \multirow{4}{*}{\underline{0.059} / \underline{-0.5700}} & \multirow{4}{*}{\underline{0.033} / \underline{-0.4957}} \\ 
&Male & 0.773 & 0.706 & 0.729 \\ 
&Avg.↑ & 0.775 & 0.694 & 0.723 \\ 
&Diff.↓ & 0.006 & 0.023 & 0.014 \\

\midrule
\multirow{4}{*}{DomainIndep~\cite{wang2020towards}} 
&Female & 0.729 & 0.747 & 0.734  & \multirow{4}{*}{0.010 / 0.8106} & \multirow{4}{*}{0.086 / -0.9597} & \multirow{4}{*}{0.042 / -0.9061} \\ 
&Male & 0.725 & 0.694 & 0.702 \\ 
&Avg.↑ & 0.727 & 0.721 & 0.718 \\ 
&Diff.↓ & 0.004 & 0.053 & 0.031 \\





\midrule
&Female & 0.787 & 0.701 & 0.736  & \multirow{4}{*}{\textbf{0.006} / \textbf{0.9018}} & \multirow{4}{*}{\textbf{0.015 / 0.6724}} & \multirow{4}{*}{\textbf{0.006 / 0.7411}} \\ 
SCP-FairPrune (Ours) &Male & 0.765 & 0.712 & 0.735 \\ 
($pr_c=2\%, n=3$) &Avg.↑ & 0.776 & 0.707 & 0.736 \\ 
&Diff.↓ & 0.022 & 0.012	& 0.001 \\

\bottomrule
\end{tabular}}
\end{center}
\end{table*}

\renewcommand{\arraystretch}{2}
\begin{table*}[!ht]
\caption{Accuracy and fairness of classification results across different baselines with and without the SNNL-based Channel Pruning framework on the Fitzpatrick17k
dataset. SCP-``X'' refers to applying our framework to the ``X'' model. ``X'' model is also the baseline used in \textit{FATE} metric evaluation. Our framework always achieves positive \textit{FATE} suggesting better accuracy-fairness trade-off. ($n$ is the pruning iteration(s), and $pr_c$ is the channel pruning ratio.) }\label{fitz_different}
\begin{center}
\renewcommand{\arraystretch}{0.8}
\resizebox{\linewidth}{!}{
\begin{tabular}{ccccccccc}
\toprule
& & \multicolumn{3}{c}{Accuracy} & \multicolumn{3}{c}{Fairness}\\
\cmidrule{3-5}\cmidrule(l{2pt}r{2pt}){6-8}
Method & Skin Tone & Precision & Recall & F1-score & Eopp0↓ / FATE↑ & Eopp1↓ / FATE↑ & Eodd↓ / FATE↑ \\


\midrule
\multirow{4}{*}{VGG-11~\cite{simonyan2014very}} 
&Dark & 0.563 & 0.581 & 0.546 & \multirow{4}{*}{0.0013 / 0.0000} & \multirow{4}{*}{0.361 / 0.0000} & \multirow{4}{*}{0.182 / 0.0000} \\ 
&Light & 0.482 & 0.495 & 0.473 \\ 
&Avg.↑ & 0.523 & 0.538 & 0.510 \\ 
&Diff.↓ & 0.081 & 0.086 & 0.073 \\

\midrule
&Dark & 0.580 & 0.583 & 0.552 & \multirow{4}{*}{\textbf{0.0013 / 0.0301}} & \multirow{4}{*}{\textbf{0.286 / 0.2371}} & \multirow{4}{*}{\textbf{0.143 / 0.2433}} \\ 
SCP-VGG-11 &Light & 0.511 & 0.506 & 0.498 \\ 
($pr_c=2\%, n=3$) &Avg.↑ & 0.545 & 0.544 & 0.525 \\ 
&Diff.↓ & 0.069 & 0.077 & 0.054 \\

\midrule
\midrule

\multirow{4}{*}{HSIC~\cite{quadrianto2019discovering}} 
&Dark & 0.548 & 0.522 & 0.513  & \multirow{4}{*}{0.0013 / 0.0000} & \multirow{4}{*}{0.331 / 0.0000} & \multirow{4}{*}{0.166 / 0.0000} \\ 
&Light & 0.513 & 0.506 & 0.486 \\ 
&Avg.↑ & 0.530 & 0.515 & 0.500 \\ 
&Diff.↓ & 0.040 & 0.018 & 0.029 \\

\midrule
&Dark & 0.525 & 0.518 & 0.504  & \multirow{4}{*}{\textbf{0.0012 / 0.0609}} & \multirow{4}{*}{\textbf{0.304 / 0.0656}} & \multirow{4}{*}{\textbf{0.152 / 0.0683}} \\ 
SCP-HSIC &Light & 0.477 & 0.510 & 0.479 \\ 
($pr_c=2\%, n=3$) &Avg.↑ & 0.501 & 0.514 & 0.492 \\ 
&Diff.↓ & 0.048 & 0.008 & 0.025 \\



\bottomrule
\end{tabular}}
\end{center}
\end{table*}

\end{document}